\documentclass[a4paper,twoside]{article}

\usepackage[nolist]{acronym}
\usepackage[hidelinks]{hyperref}
\usepackage{multirow}
\usepackage{stfloats}
\usepackage{float}
\usepackage{epsfig}
\usepackage{subcaption}
\usepackage{calc}
\usepackage{amssymb}
\usepackage{amstext}
\usepackage{amsmath}
\usepackage{amsthm}
\usepackage{multicol}
\usepackage{pslatex}
\usepackage{apalike}
\usepackage[bottom]{footmisc}
\usepackage{SCITEPRESS}     

\begin{acronym}
\acro{nlp}[NLP]{Natural Language Processing}
\acro{pos}[PoS]{Part of Speech}
\acro{plm}[PLM]{Pretrained Language Model}
\acroplural{plm}[PLMs]{Pretrained Language Models}
\end{acronym}

\begin{document}

\title{PatternRank: Leveraging Pretrained Language Models and Part of Speech for Unsupervised Keyphrase Extraction}

\author{\authorname{Tim Schopf, Simon Klimek, and Florian Matthes}
\affiliation{Department of Computer Science, Technical University of Munich, Boltzmannstrasse 3, Garching, Germany}
\email{\{tim.schopf, simon.klimek, matthes\}@tum.de}
}


\abstract{Keyphrase extraction is the process of automatically selecting a small set of most relevant phrases from a given text. Supervised keyphrase extraction approaches need large amounts of labeled training data and perform poorly outside the domain of the training data \cite{bennani-smires-etal-2018-simple}. In this paper, we present PatternRank, which leverages pretrained language models and part-of-speech for unsupervised keyphrase extraction from single documents. Our experiments show PatternRank achieves higher precision, recall and $\text{F}_{1}\text{-scores}$ than previous state-of-the-art approaches. In addition, we present the \textit{KeyphraseVectorizers}\footnote{\href{https://github.com/TimSchopf/KeyphraseVectorizers}{https://github.com/TimSchopf/KeyphraseVectorizers}} package, which allows easy modification of part-of-speech patterns for candidate keyphrase selection, and hence adaptation of our approach to any domain.}
\onecolumn \maketitle \normalsize \setcounter{footnote}{0} \vfill

\section{\uppercase{Introduction}}
\label{sec:introduction}

To quickly get an overview of the content of a text, we can use keyphrases that concisely reflect its semantic context. Keyphrases describe the most essential aspect of a text. Unlike simple keywords, keyphrases do not consist solely of single words, but of several compound words. Therefore, keyphrases provide more information about the content of a text compared to simple keywords. Supervised keyphrase extraction approaches usually achieve higher accuracy than unsupervised ones \cite{kim_medelyan_kan_baldwin_2012,caragea-etal-2014-citation,meng-etal-2017-deep}. However, supervised approaches require manually labeled training data, which often causes subjectivity issues as well as significant investment of time and money \cite{Papagiannopoulou2020ARO}. In contrast, unsupervised keyphrase extraction approaches do not have these issues and are moreover mostly domain-independent. 

Keyphrases and their vector representations are very versatile and can be used in a variety of different \ac{nlp} downstream tasks \cite{10.1145/3460824.3460826,10.1145/3535782.3535835}. For example, they can be used as features or input for document clustering and classification \cite{hulth-megyesi-2006-study,webist21}, they can support extractive summarization \cite{zhang_zincir-heywood_milios_2004}, or they can be used for query expansion \cite{10.1145/1141753.1141800}. Keyphrase extraction is particularly relevant for the scholarly domain as it helps to recommend articles, highlight missing citations to authors, identify potential reviewers for submissions, analyze research trends over time, and can be used in many different search scenarios \cite{augenstein-etal-2017-semeval}.

\begin{figure*}[ht!]
    \centering
    \includegraphics[width=\textwidth]{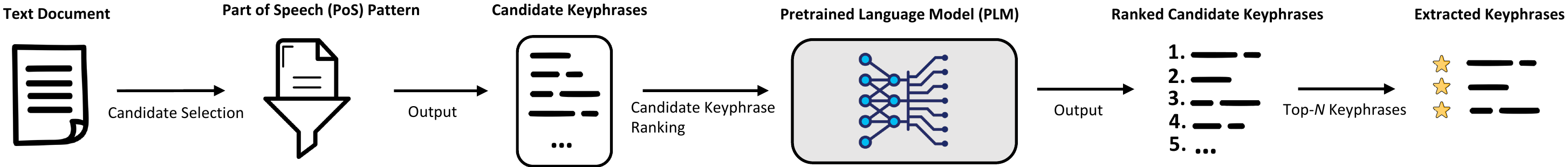}
    \caption{PatternRank approach for unsupervised keyphrase extraction. A single text document is used as input for an initial filtering step where candidate keyphrases are selected which match a defined \ac{pos} pattern. Subsequently, the candidate keyphrases are ranked by a \ac{plm} based on their semantic similarity to the input text document. Finally, the top-$N$ keyphrases are extracted as a concise reflection of the input text document.}
    \label{fig:keyphrase_extraction_pipeline}
\end{figure*}

In this paper, we present PatternRank, an unsupervised approach for keyphrase extraction based on \acp{plm} and \ac{pos}. Since keyphrase extraction is especially important for the scholarly domain, we evaluate PatternRank on a specific dataset from this area. Our approach does not rely on labeled data and therefore can be easily adapted to a variety of different domains. Moreover, PatternRank does not require the input document to be part of a larger corpus, allowing the keyphrase extraction to be applied to individual short texts such as publication abstracts. Figure \ref{fig:keyphrase_extraction_pipeline} illustrates the general keyphrase extraction approach of PatternRank. 

\section{\uppercase{Related Work}}
\label{sec:related_work}

Most popular unsupervised keyphrase extraction approaches can be characterized as either statistics-based, graph-based, or embedding-based methods, while \textit{Tf-Idf} is a common baseline used for evaluation \cite{Papagiannopoulou2020ARO}. 

\textit{YAKE} uses a set of different statistical metrics including word casing, word position, word frequency, and more to extract keyphrases from text \cite{CAMPOS2020257}. \textit{TextRank} uses \ac{pos} filters to extract noun phrase candidates that are added to a graph as nodes, while adding an edge between nodes if the words co-occur within a defined window \cite{mihalcea-tarau-2004-textrank}. Finally, PageRank \cite{Page1999ThePC} is applied to extract keyphrases. \textit{SingleRank} expands the TextRank approach by adding weights to edges based on word co-occurrences \cite{wan-xiao-2008-collabrank}. \textit{RAKE} generates a word co-occurrence graph and assigns scores based on word frequency, word degree, or the ratio of degree and frequency for keyphrase extraction \cite{Rose2010AutomaticKE}. Furthermore, Knowledge Graphs can be used to incorporate semantics for keyphrase extraction \cite{Shi2017KeyphraseEU}. \textit{EmbedRank} leverages Doc2Vec \cite{pmlr-v32-le14} and Sent2Vec \cite{pagliardini-etal-2018-unsupervised} sentence embeddings to rank candidate keyphrases for extraction \cite{bennani-smires-etal-2018-simple}. More recently, a \ac{plm}-based approach was introduced that uses BERT \cite{devlin-etal-2019-bert} for self-labeling of keyphrases and subsequent use of the generated labels in an LSTM classifier \cite{sharma2019}.

\section{\uppercase{Keyphrase Extraction Approach}}
\label{sec:approach}

Figure \ref{fig:keyphrase_extraction_pipeline} illustrates the general keyphrase extraction process of our PatternRank approach. The input consists of a single text document which is being word tokenized. The word tokens are then tagged with \ac{pos} tags. Tokens whose tags match a previously defined \ac{pos} pattern are selected as candidate keyphrases. Then, the candidate keyphrases are fed into a \ac{plm} to rank them based on their similarity to the input text document. The \ac{plm} embeds the entire text document as well as all candidate keywords as semantic vector representations. Subsequently, the cosine similarities between the document representation and the candidate keyphrase representations are computed and the candidate keyphrases are ranked in descending order based on the computed similarity scores. Finally, the top-$N$ ranked keyphrases, which are most representative of the input document, are extracted.

\subsection{Candidate Selection with Part of Speech}
\label{sec:candidate_selection}

In previous work, simple noun phrases consisting of zero or more adjectives followed by one or more nouns were used for keyphrase extraction \cite{mihalcea-tarau-2004-textrank,wan-xiao-2008-collabrank,bennani-smires-etal-2018-simple}. However, we define a more complex \ac{pos} pattern to extract candidate keyphrases from the input text document. In our approach, the tags of the word tokens have to match the following \ac{pos} pattern in order for the tokens to be considered as candidate keyphrases:

\begin{equation}
    \begin{aligned}
    &\Big(\big(\{.*\}\{HYPH\}\{.*\}\big)\{NOUN\}*\Big) \Big\vert \\ 
    &\Big(\big(\{VBG\} \vert \{VBN\}\big)?\{ADJ\}*\{NOUN\}+\Big)
    \end{aligned}
\end{equation}

The \ac{pos} pattern quantifiers correspond to the regular expression syntax. Therefore, we can translate the \ac{pos} pattern as \textit{arbitrary parts-of-speech separated by a hyphen, followed by zero or more nouns OR zero or one verb (gerund or present or past participle), followed by zero or more adjectives, followed by one or more nouns}.

\subsection{Candidate Ranking with Pretrained Language Models}
\label{sec:candidate_ranking}

Earlier work used graphs \cite{mihalcea-tarau-2004-textrank,wan-xiao-2008-collabrank} or paragraph and sentence embeddings \cite{bennani-smires-etal-2018-simple} to rank candidate keyphrases. However, we leverage \acp{plm} based on current transformer architectures to rank the candidate keyphrases that have recently demonstrated promising results \cite{grootendorst2020keybert}. Therefore, we follow the general EmbedRank \cite{bennani-smires-etal-2018-simple} approach for ranking, but use \acp{plm} instead of Doc2Vec \cite{pmlr-v32-le14} and Sent2Vec \cite{pagliardini-etal-2018-unsupervised} to create semantic vector representations of the entire text document as well as all candidate keyphrases. In our experiments, we use SBERT \cite{reimers-gurevych-2019-sentence} \acp{plm} since they have been shown to produce state of the art text representations for semantic similarity tasks. Using these semantic vector representations, we rank the candidate keyphrases based on their cosine similarity to the input text document.

\section{\uppercase{Experiments}}
\label{sec:experiments}

In this section, we compare four different approaches for unsupervised keyphrase extraction in the scholarly domain.

\subsection{{Data}}
\label{sec:data}

In our experiments, we use the \textit{Inspec} dataset \cite{10.3115/1119355.1119383}, which consists of 2,000 English computer science abstracts collected from scientific journal articles between 1998 and 2002. Each abstract has assigned two different types of keyphrases. First, controlled and manually assigned keyphrases that appear in the thesaurus of the \textit{Inspec} dataset but do not necessarily have to appear in the abstract. Second, uncontrolled keyphrases that are freely assigned by professional indexers and are not restricted to either the thesaurus or the abstract. In our experiments, we consider the union of both types of keyphrases as the ground truth.

\subsection{{Evaluation}}
\label{sec:evaluation}

For evaluation, we compare the performances of four different keyphrase extraction approaches.\\ \\ 
\textbf{YAKE} is a fast and lightweight approach for unsupervised keyphrase extraction from single documents based on statistical features \cite{CAMPOS2020257}. \\ \\
\textbf{SingleRank} applies a ranking algorithm to word co-occurrence graphs for unsupervised keyphrase extraction from single documents \cite{wan-xiao-2008-collabrank}. \\ \\
\textbf{KeyBERT} uses, similar to PatternRank, a \ac{plm} to rank candidate keyphrases \cite{grootendorst2020keybert}. However, KeyBERT uses simple word n-grams as candidate keyphrases rather than word tokens that match a certain \ac{pos} pattern, as in our PatternRank approach. For the KeyBERT experiments, we use the \textit{all-mpnet-base-v2}\footnote{\href{https://huggingface.co/sentence-transformers/all-mpnet-base-v2}{https://huggingface.co/sentence-transformers/all-mpnet-base-v2}} SBERT model for candidate keyphrase ranking and an n-gram range of $[1,3]$ for candidate keyphrase selection. This means that n-grams consisting of 1, 2 or 3 words are selected as candidate keyphrases. \\ \\
\textbf{PatternRank} To select candidate keyphrases, we developed the \textit{KeyphraseVectorizers}\footnote{\href{https://github.com/TimSchopf/KeyphraseVectorizers}{https://github.com/TimSchopf/KeyphraseVectorizers}} package, which allows custom \ac{pos} patterns to be defined and returns matching candidate keyphrases. We evaluate two different versions of the PatternRank approach. $\text{PatternRank}_{NP}$ selects simple noun phrases as candidate keyphrases and $\text{PatternRank}_{PoS}$ selects word tokens whose \ac{pos} tags match the pattern defined in section \ref{sec:candidate_selection}. In both cases, the \textit{all-mpnet-base-v2} SBERT model is used for candidate keyphrase ranking. \\ \\
We evaluate the models based on exact match, partial match, and the average of exact and partial match. For each approach, we report $\text{Precision}\text{@N}$, $\text{Recall}\text{@N}$, and $\text{F}_{1}\text{@N}$ scores, using the top-N extracted keyphrases respectively. The gold keyphrases always remain the entire set of all manually assigned keyphrases, regardless of N. Additionally, we lowercase the gold keyphrases as well as the extracted keyphrases and remove duplicates. We follow the approach of Rousseau and Vazirgiannis (2015) and calculate $\text{Precision}\text{@N}$, $\text{Recall}\text{@N}$, and $\text{F}_{1}\text{@N}$ scores per document and then use the macro-average at the collection level for evaluation. The exact match approach yields true positives only for extracted keyphrases that have an exact string match to one of the gold keyphrases. However, this evaluation approach penalizes keyphrase extraction methods which predict keyphrases that are syntactically different from the gold keyphrases but semantically similar \cite{10.1007/978-3-319-16354-3_42,10.1007/978-3-319-19548-3_21}. The partial match approach converts gold keyphrases as well as extracted keyphrases to unigrams and yields true positives if the extracted unigram keyphrases have a string match to one of the unigram gold keyphrases \cite{10.1007/978-3-319-16354-3_42}. The drawback of the partial match evaluation approach, however, is that it rewards methods which predict keyphrases that occur in the unigram gold keyphrases but are not appropriate for the corresponding document \cite{Papagiannopoulou2020ARO}. For empirical comparison of keyphrase extraction approaches, we therefore also report the average of the exact and partial matching results.

\begin{table*}[hbt!]
    \centering
    \begin{tabular}{l|l|ccc|ccc|ccc|}
    \cline{2-11}
                & \multirow{2}{*}{Method}   &   & @5 &    &   & @10 &    &   & @20 &    \\
                &                           & P & R  & $\text{F}_{1}$ & P & R   & $\text{F}_{1}$ & P & R   & $\text{F}_{1}$ \\ \hline
         \multirow{5}{*}{\rotatebox[origin=c]{90}{Exact Match}} & YAKE                      & 26.16 & 11.71  & 15.37  & 20.88 & 18.45   & 18.50  & 16.45 & 27.78   & 19.65  \\
                & SingleRank                & 38.11 & 16.55  & 21.97  & 33.29 & 27.27   & 28.55  & 27.24 & 38.84  & 30.80  \\
                & KeyBERT                    & 12.97 & 6.08   & 7.82 &  11.42 & 10.53  & 10.30 & 9.75 & 17.14 & 11.76  \\
                & $\text{PatternRank}_{NP}$            & 41.15 & 18.09  & 23.92  & 34.60 & 28.33  & 29.66  & 25.88 & 36.69 & 29.19  \\
                & $\text{PatternRank}_{PoS}$            & \textbf{41.76} & \textbf{18.44}  & \textbf{24.35}  & \textbf{36.10} & \textbf{29.63}   & \textbf{30.99}  & \textbf{27.80} & \textbf{39.42}  & \textbf{31.37}  \\ 
                \hline
         \multirow{5}{*}{\rotatebox[origin=c]{90}{Partial Match}} & YAKE                       & 77.45 & 19.49   & 29.91  & 68.20 & 33.46   & 42.67  & 59.69 & 45.58 & 48.69  \\
                & SingleRank                & 75.54 & 19.36  & 29.56  & 68.63 & 33.98  & 43.24  & 58.82 & 53.68   & 53.68  \\
                & KeyBERT                   & 77.48 & 20.06  & 30.55  & 65.78 & 32.90 &  41.67  & 57.11 & 45.37 & 48.34  \\
                & $\text{PatternRank}_{NP}$            & \textbf{83.64} & \textbf{21.93}  & \textbf{33.29}  & \textbf{75.27} & \textbf{37.62} & \textbf{47.69}  & 62.78 & 56.69 & 57.03  \\
                & $\text{PatternRank}_{PoS}$           & 82.49 & 21.61  & 32.79  & 74.79 & 37.50   & 47.48  & \textbf{63.21} & \textbf{57.66} & \textbf{57.71}  \\
                \hline
         \multirow{5}{*}{\rotatebox[origin=c]{90}{Avg. Match}} & YAKE                      & 51.81 & 15.60  & 22.64  & 44.54 & 25.96  & 30.59  & 38.07 & 36.68   & 34.13  \\
                & SingleRank                & 56.83 & 17.96  & 25.77  & 50.96 & 30.63   & 35.90  & 43.03 & 46.26   & 42.24  \\
                & KeyBERT                   & 45.23 & 13.07  & 19.19 & 38.60 & 21.72 & 25.99  & 33.43 & 31.23 & 30.05  \\
                & $\text{PatternRank}_{NP}$            & \textbf{62.40} & 20.01  & \textbf{28.61}  & 54.94 & 32.98  & 38.68  & 44.33 & 46.69   & 43.11  \\
                & $\text{PatternRank}_{PoS}$           & 62.13 & \textbf{20.03}  & 28.57  & \textbf{55.45} & \textbf{33.57}   & \textbf{39.24}  & \textbf{45.51} & \textbf{48.54}   & \textbf{44.54}  \\ \hline
    \end{tabular}
    \caption{Evaluation of our approach against state of the art using the \textit{Inspec} dataset. Precision (P), Recall (R), and $\text{F}_{1}$-score ($\text{F}_{1}$) for N = 5, 10, and 20 are reported. The evaluation results are based on exact match, partial match, and the average of exact and partial match. Two variations of PatternRank are presented: $\text{PatternRank}_{NP}$ selects simple noun phrases as candidate keyphrases and $\text{PatternRank}_{PoS}$ selects word tokens whose \ac{pos} tags match the pattern defined in section \ref{sec:candidate_selection}. In both cases, a SBERT \ac{plm} is used for candidate keyphrase ranking.}
    \label{tab:evaluation_results}
\end{table*}

The results of our evaluation are shown in Table \ref{tab:evaluation_results}. We can see that our PatternRank approach outperforms all other approaches across all benchmarks. In general, both approaches $\text{PatternRank}_{NP}$ and $\text{PatternRank}_{PoS}$ perform fairly similarly, whereas $\text{PatternRank}_{PoS}$ produces slightly better results in most cases. In the exact match evaluation, $\text{PatternRank}_{PoS}$ consistently achieves the best results of all approaches. Furthermore, $\text{PatternRank}_{PoS}$ also yields best results in the average mach evaluation for $\text{N = 10 and 20}$. In the partial match evaluation, the $\text{PatternRank}_{NP}$ approach marginally outperforms the $\text{PatternRank}_{PoS}$ approach and yields best results for $\text{N = 5 and 10}$. However, as we mentioned earlier the partial match evaluation approach, may wrongly reward methods which extract keyphrases that occur in the unigram gold keyphrases but are not appropriate for the corresponding document. Since the $\text{PatternRank}_{PoS}$ approach outperforms the $\text{PatternRank}_{NP}$ approach in the more important exact match and average match evaluations, we argue that selecting candidate keyphrases based on the \ac{pos} pattern defined in Section \ref{sec:candidate_selection} instead of simple noun phrases helps to extract keyphrases predominantly occurring in the scholarly domain. In contrast, skipping the \ac{pos} pattern-based candidate keyphrase selection step results in a significant performance decline. KeyBERT uses the same \ac{plm} to rank the candidate keyphrases as PatternRank, but uses simple n-grams for candidate keyphrase selection instead of \ac{pos} patterns or noun phrases. As a result, the KeyBERT approach consistently performs worst among all approaches. As expected, YAKE was the fastest keyphrase extraction approach because it is a lightweight method based on statistical features. However, the extracted keyphrases are not very accurate and in comparison to PatternRank, YAKE significantly performs worse in all evaluations. SingleRank is the only approach that achieves competitive results compared to PatternRank. Nevertheless, it consistently performs a few percentage points worse than PatternRank across all evaluations. We therefore conclude that our PatternRank achieves state-of-the-art keyphrase extraction results, especially in the scholarly domain.  

\section{\uppercase{Conclusion}}
\label{sec:conclusion}

We presented the PatternRank approach which leverages \acp{plm} and \ac{pos} for unsupervised keyphrase extraction. We evaluated our approach against three different keyphrase extraction methods: one statistics-based approach, one graph-based approach and one \ac{plm}-based approach. The results show that the PatternRank approach performs best in terms of precision, recall and $\text{F}_{1}\text{-score}$ across all evaluations. Furthermore, we evaluated two different PatternRank versions. $\text{PatternRank}_{NP}$ selects simple noun phrases as candidate keyphrases and $\text{PatternRank}_{PoS}$ selects word tokens whose \ac{pos} tags match the pattern defined in Section \ref{sec:candidate_selection}. While $\text{PatternRank}_{PoS}$ produced better results in the majority of cases, $\text{PatternRank}_{NP}$ still performed very well in all benchmarks. We therefore conclude that the $\text{PatternRank}_{PoS}$ approach works particularly well in the evaluated scholarly domain. Furthermore, since the use of noun phrases as candidate keyphrases is a more general and domain-independent approach, we propose using $\text{PatternRank}_{NP}$ as a simple but effective keyphrase extraction method for arbitrary domains. Future work may investigate how the \ac{plm} and \ac{pos} pattern used in this approach can be adapted to different domains or languages.

\bibliographystyle{apalike}
{\small
\bibliography{paper}}

\end{document}